\documentclass{article}

\usepackage[preprint]{neurips_2026}

\usepackage[utf8]{inputenc} 
\usepackage[T1]{fontenc}    
\usepackage{hyperref}       
\usepackage{url}            
\usepackage{booktabs}       
\usepackage{amsfonts}       
\usepackage{nicefrac}       
\usepackage{microtype}      
\usepackage{xcolor}         
\usepackage{graphicx}
\usepackage{float}
\usepackage{amsmath}
\usepackage{subcaption}

\title{Positional Encodings Anchor Spatial Structure in Vision Transformers: A Geometric Perspective on Robustness}

\author{%
  Mahmoud Mannes \\
  Undergrad Student\\
  ESSTHS\\
  \texttt{mannesmahmoud@gmail.com} \\
}

\begin{document}

\maketitle

\begin{abstract}
Positional embeddings (PEs) in Vision Transformers (ViTs) are known to impact performance and robustness, but their role in shaping internal spatial representations is not well understood. In this work, we study how different forms of PEs influence the representational geometry of ViTs and how these changes relate to robustness under content-disrupting distribution shifts.
We introduce a metric, the Spatial Similarity Distance Correlation (SSDC), to quantify spatial structure in token representations. Using this metric, we show that ViTs trained without PEs still develop non-trivial spatial structure, but this structure is driven by visual content and collapses under token permutation.
In contrast, we find that all PEs considered (learned absolute, sinusoidal, and rotary) are associated with a consistent shift toward an index-anchored spatial organization. Representations in these models remain stable under perturbations that disrupt content, and exhibit substantially improved robustness to such distributional shifts.
We further show that while different PEs produce distinct depth-wise trajectories of spatial structure, their robustness properties are largely similar (with secondary variation across encoding schemes), suggesting that robustness appears to depend on the presence of a stable positional reference frame more than it depends on the specific encoding mechanism.
These results offer a geometric account of how positional encodings shape internal representations, with implications for the principled design of future encoding schemes.
\end{abstract}

\section{Introduction}
Vision Transformers (ViTs) model images as sequences of patch tokens processed by self-attention \citep{dosovitskiy2021an}. Unlike convolutional architectures, they lack built-in inductive biases toward locality and translation equivariance, and instead rely on positional embeddings (PEs) to inject spatial information, enabling the model to distinguish tokens originating from different locations.

While PEs are designed to provide positional information, this design does not determine how that signal is integrated into internal representations. In particular, it remains unclear whether positional information organizes token representations into similarity structures anchored to absolute indices, or whether spatial structure continues to arise primarily from visual content.

Prior work shows that ViTs retain substantial performance even when positional information is removed or degraded \citep{dosovitskiy2021an, chu2023conditional}, suggesting that spatial relationships can partially emerge from patch content alone. This raises a central question: if spatial structure can arise without explicit positional guidance, what functional role do positional embeddings play?

Existing studies have largely addressed this question through downstream performance comparisons or architectural variations. While informative, these approaches provide limited insight into how positional information shapes internal representations. In particular, it remains unclear whether different positional encoding schemes (learned absolute, sinusoidal, or rotary) induce distinct spatial reasoning strategies, or whether their effects on robustness arise from a shared mechanism.

In this work, we adopt a geometric perspective. We analyze the evolution of token representations across the transformer stack using tools from representational geometry \citep{raghu2021do}, introducing the Spatial Similarity Distance Correlation (SSDC) as a probe of spatial structure. Critically, we use SSDC in conjunction with a random permutation intervention at inference to distinguish whether spatial organization is anchored to token indices or driven by patch content. We compare models trained with learned absolute positional embeddings (APE), sinusoidal encodings (SPE), rotary embeddings (RoPE), and no positional embeddings, and evaluate their robustness to distributional shifts.

Our central finding is that the specific encoding mechanism matters less than the presence of a consistent positional signal. We show that:
\begin{itemize}
    \item \textbf{Positional encodings are associated with index-based spatial organization:} All PE types shift ViTs away from purely content-driven spatial structure toward representations that remain partially anchored to token indices under permutation.
    \item \textbf{This shift, not the encoding form, is associated with robustness:} Despite differing in how spatial structure develops across depth, APE, sinusoidal, and RoPE models exhibit broadly comparable robustness to content-disrupting distributional shifts (despite consistent but smaller differences between encoding schemes), while models lacking index-based organization are substantially more fragile.
    \item \textbf{A stable positional reference frame is strongly implicated in robustness:} Using Random Permutation Training (RPT), which preserves PEs but destroys index-to-location consistency, we find that robustness is greatly reduced when a consistent positional frame cannot be learned.
\end{itemize}

Together, these results provide a unified, geometric account of how positional encodings shape internal representations and why they remain critical for robust visual recognition, though we emphasize that the evidence is intervention-based rather than strictly causal.
    
\section{Related Work}

\textbf{Positional Information in Vision Transformers}

The standard Vision Transformer (ViT) breaks the permutation invariance of self-attention by adding learnable absolute positional embeddings (PEs) to patch tokens \citep{dosovitskiy2021an}, establishing the dominant paradigm for spatial encoding. However, ViTs retain substantial performance when positional information is degraded or removed \citep{dosovitskiy2021an, chu2023conditional}, suggesting that spatial structure can partially emerge from patch content alone.

Similar observations have been reported beyond vision. Recent work on decoder-only transformers shows that models trained without PEs can recover positional information implicitly and tend to rely on relative positions in practice \citep{kazemnejad2023the}. Earlier findings in convolutional networks further demonstrate that substantial positional information can be learned implicitly from architectural biases such as zero-padding \citep{Islam*2020How}. Together, these results suggest that explicit positional signals are not strictly required for structured spatial information to emerge.

This creates a central puzzle: if spatial structure can arise without explicit positional guidance, what functional role do PEs play? Prior work has primarily addressed this question through architectural variants \citep{d’Ascoli_2022,Liu2021swin,Heo2024RoPE} or performance comparisons \citep{dosovitskiy2021an, chu2023conditional}, leaving their mechanistic impact on internal representations largely unexplored.

\textbf{Representational Analysis of Transformers}

A separate line of work studies the geometry and dynamics of transformer representations. Early analyses compare ViT and CNN representations \citep{raghu2021do}, revealing differences in spatial organization. Subsequent work examines how attention transforms representations \citep{kobayashi-etal-2021-incorporating}, how representational rank evolves with depth \citep{pmlr-v139-dong21a}, and how token representations tend to homogenize in deeper layers \citep{Bhojanapalli2021robust}. The residual stream framework provides a useful lens for analyzing these dynamics \citep{elhage2021mathematical}. However, these approaches do not isolate the causal role of positional embeddings, nor do they connect representational structure to robustness.

\textbf{Robustness of Visual Models}

Vision Transformers exhibit distinct robustness profiles compared to convolutional networks. Prior work shows that transformers are generally more robust to certain spatial perturbations but can be more sensitive to texture-based changes \citep{Bhojanapalli2021robust}. Additional studies report favorable out-of-distribution generalization properties for ViTs \citep{Paul_Chen_2022}, connecting to broader findings on shape versus texture bias in visual recognition \citep{geirhos2018imagenettrained}. While the impact of positional embeddings on robustness has been observed (particularly that models trained with PEs exhibit better robustness profiles than models trained without them) \citep{Mao2021TowardsRV}, the relationship between a model’s spatial organization strategy (whether anchored to absolute position or inferred from content) and its robustness to distributional shifts remains poorly understood.

\textbf{Our Contribution}

We connect these lines of work by showing that positional embeddings are associated with a shift toward index-based spatial organization, and that this shift (rather than the specific encoding mechanism) appears to be a dominant correlate of robustness. Using SSDC and controlled permutation interventions (RPT and RPI), we provide a geometric account of how positional information shapes internal representations and why it improves robustness.

\section{Preliminaries}

\subsection{Vision Transformer Architecture and Positional Encodings}

All models are Vision Transformers trained from scratch on ImageNet-100 (a subset of Imagenet-1K) \citep{5206848}, with approximately 22M parameters (details in Appendix A). Images are partitioned into fixed-size patches, projected into token embeddings, and processed by a stack of self-attention and feedforward layers.

Since self-attention is permutation invariant, positional encodings are required to inject spatial information. We consider three commonly used PE schemes, all adapted to 2D grids:

\textbf{Learned Absolute Positional Embeddings (APE):} learnable vectors added to token embeddings before the first transformer block, establishing a fixed index-to-location mapping.

\textbf{Sinusoidal Positional Embeddings (SPE):} fixed, deterministic encodings constructed from sinusoidal functions applied independently along spatial axes and added to token embeddings.

\textbf{Rotary Positional Embeddings (RoPE):} position-dependent rotations applied to query and key vectors within each attention layer, introducing positional information multiplicatively.

These approaches differ in parameterization (learned vs.\ fixed) and integration (additive vs.\ multiplicative), enabling comparison of how different positional signals shape internal representations.

\subsection{Index-Based and Content-Based Spatial Organization}

We distinguish between two qualitatively distinct modes of spatial organization.

\textbf{Index-based spatial organization} refers to representations whose similarity structure depends systematically on token position. Tokens that are spatially proximate tend to have more similar representations by virtue of their indices, and this structure persists under disruptions to patch content. This definition is behavioral and does not assume explicit coordinate representations.

\textbf{Content-based spatial organization} refers to representations in which similarity is driven primarily by patch content. Spatial structure arises indirectly from natural image statistics and degrades under transformations that disrupt content or token ordering.

In practice, models may exhibit both behaviors; the key distinction is which signal dominates.

\section{Methods}
\subsection{Residual Stream Geometry}
At selected layers, we extract the residual stream as a matrix $R \in \mathbb{R}^{T \times C}$, where $T$ is the number of tokens and $C$ the embedding dimension. We compute pairwise cosine similarities between unit-normalized token representations to form a symmetric similarity matrix, averaged across the batch dimension.

\subsection{Spatial Similarity Distance Correlation}
Let $S \in \mathbb{R}^{T \times T}$ denote the token similarity matrix, and let $p_i \in \mathbb{N}^2$ denote the spatial coordinates of token $i$. Define the spatial distance matrix $D$ by $D_{ij} = \|p_i - p_j\|_1$. We define SSDC as the Spearman rank correlation between similarity and negative spatial distance over all token pairs:
\[
\mathrm{SSDC} = \rho_{\mathrm{Spearman}}\left(\{S_{ij}\}_{i < j},\ \{-D_{ij}\}_{i < j}\right).
\]
Higher SSDC values indicate that spatially proximate tokens tend to have more similar representations. We use Spearman rank correlation to remain agnostic to the precise functional form relating spatial distance and representational similarity.

Importantly, SSDC should be interpreted as a coarse proxy for spatial organization rather than a direct measurement of a specific mechanism. Absolute values may reflect multiple factors (e.g., data statistics, architectural biases), and therefore SSDC is primarily used comparatively (to track changes across depth and to measure sensitivity to controlled interventions).
\subsection{Random Permutation at Inference (RPI)}
To distinguish index-based from content-based organization, we randomly permute token order at inference while keeping positional indices fixed. This breaks the correspondence between token order and spatial location. Under this setup, spatial structure driven purely by patch content is expected to be disrupted, as spatially adjacent tokens no longer correspond to neighboring image patches. In contrast, if a model has learned representations that depend systematically on token indices via positional signals, some spatial structure may persist or be partially recoverable.

As a result, SSDC under RPI should be interpreted as an indicator of the extent to which spatial organization depends on token indices, rather than as a definitive separation between index-based and content-based mechanisms.
\subsection{Random Permutation during Training (RPT)}
Random Permutation Training (RPT) applies a fresh random permutation to the token sequence at every forward pass during training. At each batch, patch tokens are shuffled while positional embeddings remain fixed to their original indices, breaking the consistent mapping between token index and spatial location. This prevents the model from learning a stable index-based spatial organization despite the presence of positional signals.
\subsection{Positional Embedding Magnitude Scaling}
We scale positional embeddings at inference by a factor $\alpha$, replacing $\mathbf{e}_i$ with $\alpha \mathbf{e}_i$. This provides a continuous intervention on positional signal strength without retraining. We apply this to APE and sinusoidal models; an equivalent scaling for RoPE is not directly defined due to its multiplicative formulation.
\subsection{Fragility Score}
We quantify robustness using the Fragility Score (FS):
\[
\mathrm{FS} = 1 - \frac{A_{\mathrm{shift}}}{A_{\mathrm{normal}}},
\]
where $A_{\mathrm{normal}}$ and $A_{\mathrm{shift}}$ denote accuracy on clean and shifted data. Higher values indicate greater sensitivity to distributional shift.


\section{Results}
  \subsection{Architectural Priors Induce Static Spatial Correlations at Initialization}
    \begin{figure}[t]
      \centering
      \begin{subfigure}[t]{0.47\textwidth}
        \centering
        \includegraphics[width=\textwidth]{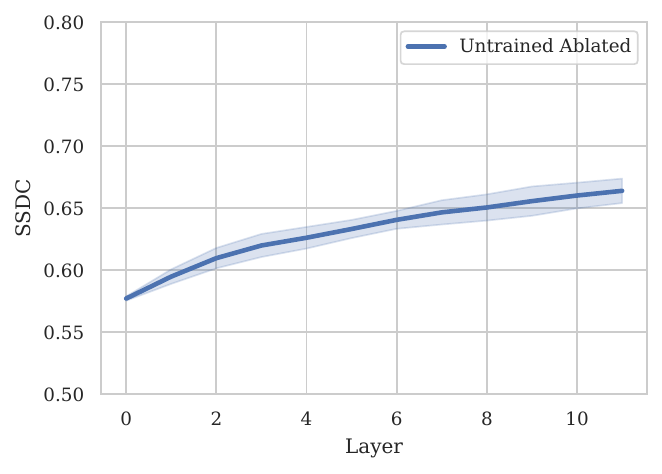}
        \caption{\textbf{Evolution of SSDC across depth on untrained ablated models}}
        \label{fig:SSDC1}
      \end{subfigure}
      \hfill
      \begin{subfigure}[t]{0.47\textwidth}
        \centering
        \includegraphics[width=\textwidth]{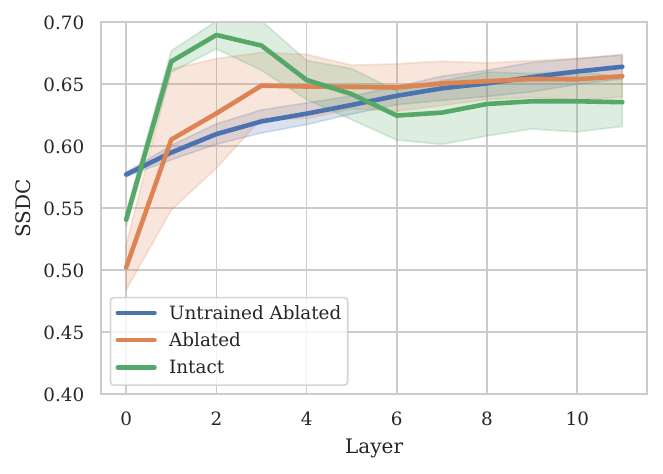}
        \caption{\textbf{Evolution of SSDC across depth on untrained ablated models, trained ablated models, and intact (trained with APE) models}}
        \label{fig:SSDC2}
      \end{subfigure}
      \caption{(a) SSDC grows weakly and remains at a relatively high value across layers, indicating static spatial correlations induced by architectural and data priors rather than learning.
      \newline (b) While untrained ablated models exhibit relatively high but slowly varying SSDC consistent with static data and architectural priors, trained ablated models display a sharp increase in early layers, indicating the emergence of learned spatial structure despite the absence of explicit positional encoding.
      }
      \label{fig:SSDC}
    \end{figure}

    \textbf{Experimental Setup:} We evaluate SSDC across all layers of untrained ablated models on the Imagenet-100 dataset. Unless stated otherwise, all reported results are averaged over 4 random seeds. Shaded regions in figures indicate variability across runs (±1 standard deviation).

    \textbf{Results:}
    The untrained ablated model exhibits a substantial non-zero SSDC (approximately 0.57–0.64) with only a weak, gradual increase across depth (Figure~\ref{fig:SSDC1}). This behavior is highly consistent across runs and reflects static spatial correlations induced by architectural priors and the inherent structure of natural images, rather than learned spatial reasoning.

    Crucially, this baseline highlights that SSDC should not be interpreted as a standalone metric whose absolute magnitude reflects the presence or strength of learned spatial organization. Even in the absence of training, relatively high SSDC values emerge. Instead, the layer-wise dynamics of SSDC (in particular, the rate and pattern of change across depth) are the informative signal. In contrast to the shallow, nearly static progression observed here, trained models exhibit rapid and structured changes in SSDC (e.g., sharp increases in early layers), indicating the emergence of learned spatial structure.

    This establishes a static baseline, allowing us to distinguish genuinely learned spatial organization from correlations that arise purely from architectural and data-driven effects.

  \subsection{Emergence of Spatial Structure Without Positional Encoding}
    \textbf{Experimental Setup:} To investigate whether spatial structure can emerge in the absence of explicit positional information, we evaluate SSDC across all layers of untrained ablated models, trained ablated models, and trained intact (APE) models on the Imagenet-100 dataset.

    \textbf{Results:}
    Figure~\ref{fig:SSDC2} compares the layer-wise evolution of SSDC for an untrained ablated model, a trained ablated model, and a trained model with positional embeddings. The untrained ablated model exhibits relatively high SSDC (approximately 0.57–0.64) with only weak growth across depth, reflecting static spatial correlations induced by architectural and data priors rather than learning.

    In contrast, the trained ablated model shows a qualitatively different trajectory: starting from lower SSDC, it exhibits a sharp increase in early layers followed by continued growth. This dynamic pattern closely resembles that of the trained model with positional embeddings. The key distinction is not absolute SSDC magnitude, but its evolution.

    These results indicate that non-trivial spatial structure emerges during training even without positional embeddings. This is consistent with the non-trivial performance of ablated models and prior evidence that transformers can implicitly recover positional information.

    We emphasize that this emergent structure is not equivalent to that induced by positional embeddings. Rather, this establishes that spatial organization can arise without explicit positional signals, motivating a more precise characterization of its underlying mechanism in the next section.
  \subsection{Disentangling Index-Based and Content-Based Spatial Organization}

    \begin{figure*}[t]
      \centering
      \begin{subfigure}[t]{0.48\textwidth}
          \centering
          \includegraphics[width=\textwidth]{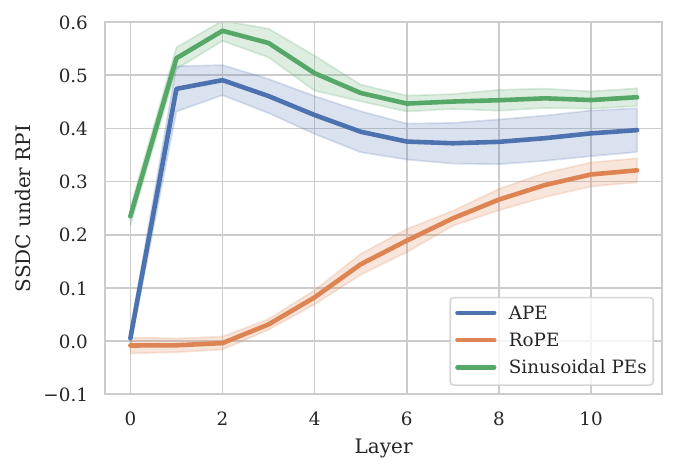}
          \caption{\textbf{Models with positional encodings.} APE, Sinusoidal PEs, and RoPE models exhibit substantial SSDC recovery under RPI, indicating spatial organization anchored to token indices. RoPE shows a gradual accumulation of structure across depth, while APE and Sinusoidal PEs exhibit earlier peaks.}
          \label{fig:ssdc_rpi_pe}
      \end{subfigure}
      \hfill
      \begin{subfigure}[t]{0.48\textwidth}
          \centering
          \includegraphics[width=\textwidth]{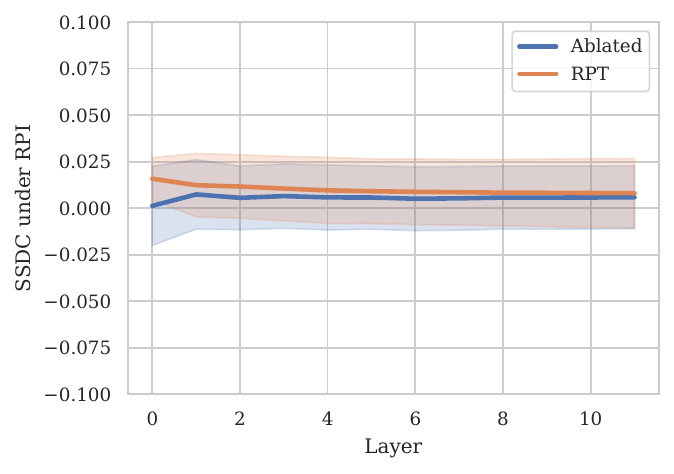}
          \caption{\textbf{Models without a stable positional reference frame.} Ablated and RPT models collapse to near-zero SSDC across all layers under RPI, indicating that their spatial structure is entirely content-driven and does not survive token permutation.}
          \label{fig:ssdc_rpi_cbso}
      \end{subfigure}
      \caption{\textbf{SSDC under random permutation at inference (RPI).} RPI disrupts the correspondence between token content and spatial position. Only models that anchor spatial structure to token indices exhibit SSDC recovery after permutation. In contrast, models lacking a consistent positional mapping collapse to near-zero SSDC, revealing a purely content-based spatial organization.}
      \label{fig:ssdc_rpi}
  \end{figure*}

    \textbf{Experimental Setup:}
    To distinguish between index-based and content-based spatial organization, we evaluate SSDC across all layers under a \emph{Random Permutation at Inference} (RPI) intervention. Concretely, patch tokens are randomly permuted before being processed by the transformer, while positional embedding indices (when present) remain fixed to their original spatial locations. This operation disrupts the correspondence between token content and spatial position, while preserving any mapping between token indices and positional embeddings.

    Under this setup, any spatial structure that arises purely from patch content is destroyed, as spatially adjacent tokens no longer correspond to neighboring image patches. In contrast, if a model has learned to anchor its representations to absolute token indices via positional embeddings, spatial structure can be re-established through the fixed positional signal. As a result, \emph{SSDC recovery under RPI} serves as a probe for index-based spatial organization: models that rely on absolute positional information exhibit non-trivial SSDC despite permutation, whereas models that rely on content-based cues collapse to near-zero SSDC.

    We evaluate this behavior across models trained with learned absolute positional embeddings (APE), sinusoidal encodings, rotary embeddings (RoPE), no positional embeddings (ablated), and under Random Permutation Training (RPT).

    \textbf{Results:}
    Models trained without positional embeddings exhibit a complete collapse of SSDC under RPI across all layers, suggesting that their spatial structure is predominantly content-driven under this probe. Despite exhibiting non-trivial SSDC in the unpermuted setting (Section~5.2), this structure does not survive disruption of patch content, indicating that it is not anchored to token indices.

    In contrast, all models trained with positional embeddings show substantial SSDC recovery under RPI, indicating representations that are more consistent with index-anchored spatial organization. However, the nature of this recovery differs across encoding schemes. For APE and sinusoidal embeddings, SSDC exhibits a rapid increase in early layers following permutation, reaching a peak within the first few layers before stabilizing or slightly decreasing. This behavior suggests that spatial structure is injected early in the network via additive positional signals.

    RoPE models display a qualitatively different trajectory: SSDC increases more gradually and continues to grow with depth, without a pronounced early-layer peak. This indicates that positional information is integrated progressively throughout the network, consistent with its multiplicative incorporation into attention mechanisms. A similar depth-wise pattern is observed in the unpermuted setting (Appendix~C.1).

    RPT models, despite having positional embeddings present, fail to exhibit meaningful SSDC recovery under RPI, behaving similarly to fully ablated models. This suggests that the mere presence of positional embeddings is insufficient; a consistent mapping between token indices and spatial locations during training appears necessary for index-based spatial organization to emerge.

    Taken together, these results establish that positional embeddings are associated with a shift from content-based to index-based spatial organization, and that this shift depends critically on the stability of the positional reference frame rather than on the architectural presence of positional signals alone.

  \subsection{Robustness to Content-Preserving and Content-Disrupting Perturbations}

    \begin{figure*}[t]
      \centering
      \begin{subfigure}[t]{0.48\textwidth}
          \centering
          \includegraphics[width=\linewidth]{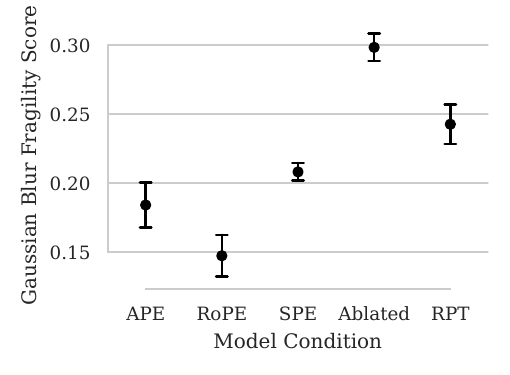}
          \caption{\textbf{Gaussian Blur ($\sigma = 2.5$).} Fragility scores under a mild perturbation that removes high-frequency detail while largely preserving global spatial structure. Differences between models are present but compressed, reflecting the weaker disruption of content-based cues.}
          \label{fig:fs_blur}
      \end{subfigure}
      \hfill
      \begin{subfigure}[t]{0.48\textwidth}
          \centering
          \includegraphics[width=\linewidth]{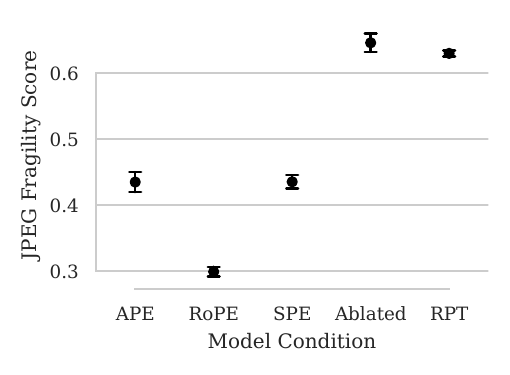}
          \caption{\textbf{JPEG Compression (quality = 5).} Fragility scores under a strong content-disrupting transformation. Models with positional encodings exhibit substantially lower fragility than ablated and RPT models, with RoPE achieving the lowest fragility overall.}
          \label{fig:fs_jpeg}
      \end{subfigure}
      \caption{\textbf{Robustness to distributional shifts.} Fragility scores across model variants under two perturbation regimes. The gap between models with and without a stable positional reference frame is most pronounced under strong content disruption (JPEG), while remaining consistent but attenuated under milder perturbations (Gaussian blur).}
      \label{fig:fragility}
    \end{figure*}
    
    \textbf{Experimental Setup:}
    To evaluate how spatial organization strategy influences robustness, we measure performance under distribution shifts that perturb image content while preserving global structure. We consider two transformations:

    \textbf{JPEG Compression:} We apply aggressive compression (quality = 5), introducing blocking artifacts that strongly disrupt local texture statistics while preserving coarse spatial layout. This provides a targeted probe of reliance on content-based cues.

    \textbf{Gaussian Blur:} We apply Gaussian blur with standard deviation $\sigma = 2.5$, attenuating high-frequency detail while preserving low-frequency structure. This constitutes a milder perturbation than JPEG.

    For each model, we compute the \emph{Fragility Score} (FS), defined as the relative drop in accuracy under each transformation. We also report the raw accuracy of each model condition in Appendix D.

    We include \textbf{Random Permutation Training (RPT)} as a critical control, allowing us to distinguish between the mere presence of positional signals and the emergence of a consistent positional reference frame.

    We emphasize that these robustness results are limited to content-disrupting perturbations (e.g., compression artifacts and blur) and do not necessarily generalize to other forms of distribution shift.
    
    \textbf{Results:}
    Under JPEG compression, models with positional encodings exhibit substantially lower fragility (APE and sinusoidal: $\sim$0.43, RoPE: $\sim$0.30) than ablated and RPT models ($\sim$0.66). This large gap indicates that robustness to severe content degradation is strongly influenced by the presence of a stable positional reference frame. Within PE-based models, RoPE consistently achieves lower fragility, suggesting a secondary effect of the encoding mechanism. We speculate that RoPE's progressive depth-wise accumulation of spatial structure may keep later layers more spatially grounded than the early-layer injection characteristic of additive encodings.

    Under Gaussian blur, the same ordering is preserved but differences are attenuated (RoPE: $\sim$0.15, APE: $\sim$0.17--0.20, sinusoidal: $\sim$0.22, RPT: $\sim$0.25, ablated: $\sim$0.30). Because blur preserves global structure, it provides a weaker test of reliance on content-based cues, reducing the separation between models.

    Taken together, these results support a two-level interpretation: (1) the emergence of a stable positional reference frame appears to be a dominant factor associated with robustness, and (2) the specific encoding mechanism introduces secondary variation, with RoPE exhibiting consistently lower fragility.
    Crucially, the poor robustness of RPT models shows that the mere presence of positional embeddings is insufficient: robustness appears to rely on learning a consistent mapping between token indices and spatial locations. This provides evidence for a relationship between the spatial organization patterns identified earlier and downstream robustness.

  \subsection{Linking Index-Based Spatial Organization to Robustness via Positional Scaling}

    \textbf{Experimental Setup:}
    To probe the relationship between spatial organization and robustness, we require a controlled intervention that selectively disrupts index-based spatial structure while preserving the rest of the model. We achieve this by scaling the magnitude of learned absolute positional embeddings (APE) at inference time by a factor $\alpha \in [0,1]$, without retraining.

    While this intervention operates on positional embeddings, our goal is not to study positional signal strength per se, but to use it as a mechanism to continuously degrade the model’s \emph{index-based spatial organization}. To measure the integrity of this organization, we evaluate Spatial Similarity Distance Correlation (SSDC) under Random Permutation at Inference (RPI), as introduced in Section~5.3. Under RPI, any recovered spatial structure must be anchored to token indices rather than content. We therefore interpret SSDC recovery as a proxy for the presence of index-based spatial organization. Importantly, SSDC recovery approaching zero does not imply content-based spatial structure; it indicates that the positional signal is too weak to sustain index-based organization.

    To summarize this behavior compactly, we define:
    \[
    \Delta \mathrm{SSDC} = \mathrm{SSDC}_{\text{layer }1} - \mathrm{SSDC}_{\text{layer }0},
    \]
    which captures the immediate recovery of spatial structure after the first encoder block under RPI. Thus, $\Delta \mathrm{SSDC}$ serves as a measure of index-based spatial organization.

    We jointly analyze $\Delta \mathrm{SSDC}$ and the Fragility Score (FS) across varying $\alpha$. For clarity, we report representative magnitudes illustrating distinct regimes, with the full results provided in Appendix B.1 (and Appendix B.2 for Sinusoidal PEs).

    \begin{table}[H]
    \centering
    \small
    \begin{tabular}{c|cc|cc}
    \toprule
    \textbf{$\alpha$} & \multicolumn{2}{c|}{\textbf{$\Delta$SSDC (RPI)}} & \multicolumn{2}{c}{\textbf{Fragility Score}} \\
    & Mean & Std & Mean & Std \\
    \midrule
    1.0 & 0.4725 & 0.0228 & 0.4338 & 0.0127 \\
    0.8 & 0.3125 & 0.0259 & 0.4780 & 0.0146 \\
    0.7 & 0.1845 & 0.0342 & 0.5145 & 0.0145 \\
    0.5 & 0.0475 & 0.0083 & 0.5975 & 0.0202 \\
    0.4 & 0.0000 & 0.0000 & 0.6272 & 0.0189 \\
    \bottomrule
    \end{tabular}
    \caption{
    Effect of positional embedding magnitude $\alpha$ on index-based spatial organization and robustness.
    $\Delta$SSDC captures the recovery of index-based spatial structure after the first encoder block.
    As $\alpha$ decreases, $\Delta$SSDC collapses, indicating the breakdown of index-based spatial organization, while fragility increases sharply in the same regime before plateauing once spatial structure is lost.
    }
    \label{tab:pe_scaling}
    \end{table}

    \textbf{Results:}
    We observe a clear correspondence between the degradation of index-based spatial organization and the loss of robustness.

    At high magnitudes ($\alpha \geq 0.9$), models exhibit strong SSDC recovery ($\Delta \mathrm{SSDC} \approx 0.37$–$0.47$), indicating intact index-based spatial organization. In this regime, fragility remains relatively low and stable (FS $\approx 0.44$–$0.46$), suggesting that robustness is preserved when spatial structure is intact.

    As $\alpha$ decreases into an intermediate regime ($0.8 \geq \alpha \geq 0.5$), SSDC recovery drops sharply ($\Delta \mathrm{SSDC} \approx 0.30 \rightarrow 0.02$), reflecting the progressive breakdown of index-based spatial organization. This degradation is accompanied by a pronounced increase in fragility (FS $\approx 0.48 \rightarrow 0.63$). Notably, the most significant increases in fragility occur precisely where SSDC recovery is actively decreasing, indicating that robustness degradation is strongly correlated with the loss of spatial structure.

    Below a critical threshold ($\alpha \leq 0.4$), SSDC recovery collapses to zero ($\Delta \mathrm{SSDC} \approx 0$), indicating that index-based spatial organization is no longer recoverable under permutation. In this regime, fragility continues to increase, but only marginally (FS $\approx 0.65 \rightarrow 0.685$). This suggests that once spatial organization is fully disrupted, further degradation in robustness is no longer correlated with changes in spatial structure, but instead reflects secondary effects such as reduced representational quality or distribution mismatch induced by scaling.

    A complementary effect is observed at high magnitudes: when index-based spatial organization is already fully intact, small reductions in $\alpha$ have limited impact on fragility. Together, these observations reveal three regimes: (1) a stable regime with intact spatial organization and low fragility, (2) a transition regime where spatial structure degrades and fragility increases sharply, and (3) a collapsed regime where spatial organization is absent and fragility plateaus.

    Overall, these results provide evidence that robustness may be driven in part by the presence of index-based spatial organization. Positional scaling serves only as a means of intervention; the observed changes in robustness track the degradation of spatial structure rather than the magnitude of the positional signal itself.

\section{Limitations}

The findings reported here are based on ViT-S models trained from scratch on ImageNet-100, and it remains an open question whether the observed relationships between positional encoding, index-based spatial organization, and robustness generalize to larger architectures, pre-trained models, or models fine-tuned from large-scale checkpoints. The robustness evaluation is specifically scoped to content-disrupting perturbations (JPEG compression and Gaussian blur); we make no claims about spatial perturbations, adversarial shifts, or semantic distribution changes, and these may involve different mechanisms. SSDC is used as a coarse proxy for spatial organization rather than a direct measurement of a specific representational mechanism, and its interpretation depends on the comparative and intervention-based framing established in Section 5.1. Finally, the positional scaling experiment (Section 5.5) conflates spatial organization degradation with changes in raw positional signal magnitude, and while the three-regime structure is consistent with a mediating role for index-based organization, alternative pathways cannot be fully excluded.

\section{Conclusion}

We studied how positional encodings shape spatial organization in Vision Transformers and its relationship to robustness under content-disrupting perturbations. Using SSDC and permutation-based interventions, we found that spatial structure emerges even without positional encodings, but remains content-driven and collapses under token permutation. Models with positional encodings exhibit representations more consistent with index-anchored spatial organization. Across experiments, robustness under content-disrupting shifts is closely associated with a stable positional reference frame rather than the mere presence of positional embeddings — evidenced by RPT models and positional scaling, where robustness degrades alongside the breakdown of index-anchored spatial structure. Differences between encoding schemes persist but appear secondary. Overall, our results suggest positional encodings contribute to robustness by supporting a stable positional reference frame, though we emphasize this conclusion is based on intervention-based evidence and identifies a strong relationship rather than a fully isolated causal mechanism.

\bibliographystyle{plainnat}
\bibliography{references}

@inproceedings{
dosovitskiy2021an,
title={An Image is Worth 16x16 Words: Transformers for Image Recognition at Scale},
author={Alexey Dosovitskiy and Lucas Beyer and Alexander Kolesnikov and Dirk Weissenborn and Xiaohua Zhai and Thomas Unterthiner and Mostafa Dehghani and Matthias Minderer and Georg Heigold and Sylvain Gelly and Jakob Uszkoreit and Neil Houlsby},
booktitle={International Conference on Learning Representations},
year={2021},
url={https://openreview.net/forum?id=YicbFdNTTy}
}

@inproceedings{
chu2023conditional,
title={Conditional Positional Encodings for Vision Transformers},
author={Xiangxiang Chu and Zhi Tian and Bo Zhang and Xinlong Wang and Chunhua Shen},
booktitle={The Eleventh International Conference on Learning Representations },
year={2023},
url={https://openreview.net/forum?id=3KWnuT-R1bh}
}

@inproceedings{
raghu2021do,
title={Do Vision Transformers See Like Convolutional Neural Networks?},
author={Maithra Raghu and Thomas Unterthiner and Simon Kornblith and Chiyuan Zhang and Alexey Dosovitskiy},
booktitle={Advances in Neural Information Processing Systems},
editor={A. Beygelzimer and Y. Dauphin and P. Liang and J. Wortman Vaughan},
year={2021},
url={https://openreview.net/forum?id=R-616EWWKF5}
}

@inproceedings{
Islam*2020How,
title={How much Position Information Do Convolutional Neural Networks Encode?},
author={Md Amirul Islam* and Sen Jia* and Neil D. B. Bruce},
booktitle={International Conference on Learning Representations},
year={2020},
url={https://openreview.net/forum?id=rJeB36NKvB}
}

@inproceedings{
kazemnejad2023the,
title={The Impact of Positional Encoding on Length Generalization in Transformers},
author={Amirhossein Kazemnejad and Inkit Padhi and Karthikeyan Natesan and Payel Das and Siva Reddy},
booktitle={Thirty-seventh Conference on Neural Information Processing Systems},
year={2023},
url={https://openreview.net/forum?id=Drrl2gcjzl}
}

@INPROCEEDINGS {Liu2021swin,
author = { Liu, Ze and Lin, Yutong and Cao, Yue and Hu, Han and Wei, Yixuan and Zhang, Zheng and Lin, Stephen and Guo, Baining },
booktitle = { 2021 IEEE/CVF International Conference on Computer Vision (ICCV) },
title = {{ Swin Transformer: Hierarchical Vision Transformer using Shifted Windows }},
year = {2021},
pages = {9992-10002},
doi = {10.1109/ICCV48922.2021.00986},
url = {https://doi.ieeecomputersociety.org/10.1109/ICCV48922.2021.00986},
publisher = {IEEE Computer Society},
month =Oct}

@article{d’Ascoli_2022,
doi = {10.1088/1742-5468/ac9830},
url = {https://doi.org/10.1088/1742-5468/ac9830},
year = {2022},
month = {nov},
publisher = {IOP Publishing and SISSA},
volume = {2022},
number = {11},
pages = {114005},
author = {d’Ascoli, Stéphane and Touvron, Hugo and Leavitt, Matthew L and Morcos, Ari S and Biroli, Giulio and Sagun, Levent},
title = {ConViT: improving vision transformers with soft convolutional inductive biases*},
journal = {Journal of Statistical Mechanics: Theory and Experiment},
}

@inproceedings{kobayashi-etal-2021-incorporating,
    title = "{I}ncorporating {R}esidual and {N}ormalization {L}ayers into {A}nalysis of {M}asked {L}anguage {M}odels",
    author = "Kobayashi, Goro  and
      Kuribayashi, Tatsuki  and
      Yokoi, Sho  and
      Inui, Kentaro",
    editor = "Moens, Marie-Francine  and
      Huang, Xuanjing  and
      Specia, Lucia  and
      Yih, Scott Wen-tau",
    booktitle = "Proceedings of the 2021 Conference on Empirical Methods in Natural Language Processing",
    month = nov,
    year = "2021",
    publisher = "Association for Computational Linguistics",
    url = "https://aclanthology.org/2021.emnlp-main.373/",
    doi = "10.18653/v1/2021.emnlp-main.373",
    pages = "4547--4568",
}

@InProceedings{pmlr-v139-dong21a,
  title = 	 {Attention is not all you need: pure attention loses rank doubly exponentially with depth},
  author =       {Dong, Yihe and Cordonnier, Jean-Baptiste and Loukas, Andreas},
  booktitle = 	 {Proceedings of the 38th International Conference on Machine Learning},
  pages = 	 {2793--2803},
  year = 	 {2021},
  editor = 	 {Meila, Marina and Zhang, Tong},
  volume = 	 {139},
  series = 	 {Proceedings of Machine Learning Research},
  month = 	 {18--24 Jul},
  publisher =    {PMLR},
  url = 	 {https://proceedings.mlr.press/v139/dong21a.html},
}

@inproceedings{Bhojanapalli2021robust,
author = {Bhojanapalli, Srinadh and Chakrabarti, Ayan and Glasner, Daniel and Li, Daliang and Unterthiner, Thomas and Veit, Andreas},
year = {2021},
month = {10},
pages = {10211-10221},
title = {Understanding Robustness of Transformers for Image Classification},
doi = {10.1109/ICCV48922.2021.01007}
}

@article{elhage2021mathematical,
   title={A Mathematical Framework for Transformer Circuits},
   author={Elhage, Nelson and Nanda, Neel and Olsson, Catherine and Henighan, Tom and Joseph, Nicholas and Mann, Ben and Askell, Amanda and Bai, Yuntao and Chen, Anna and Conerly, Tom and DasSarma, Nova and Drain, Dawn and Ganguli, Deep and Hatfield-Dodds, Zac and Hernandez, Danny and Jones, Andy and Kernion, Jackson and Lovitt, Liane and Ndousse, Kamal and Amodei, Dario and Brown, Tom and Clark, Jack and Kaplan, Jared and McCandlish, Sam and Olah, Chris},
   year={2021},
   journal={Transformer Circuits Thread},
   note={https://transformer-circuits.pub/2021/framework/index.html}
}

@article{Paul_Chen_2022, title={Vision Transformers Are Robust Learners}, volume={36}, url={https://ojs.aaai.org/index.php/AAAI/article/view/20103}, DOI={10.1609/aaai.v36i2.20103}, number={2}, journal={Proceedings of the AAAI Conference on Artificial Intelligence}, author={Paul, Sayak and Chen, Pin-Yu}, year={2022}, month={Jun.}, pages={2071-2081} }

@inproceedings{
geirhos2018imagenettrained,
title={ImageNet-trained {CNN}s are biased towards texture; increasing shape bias improves accuracy and robustness.},
author={Robert Geirhos and Patricia Rubisch and Claudio Michaelis and Matthias Bethge and Felix A. Wichmann and Wieland Brendel},
booktitle={International Conference on Learning Representations},
year={2019},
url={https://openreview.net/forum?id=Bygh9j09KX},
}

@inproceedings{Heo2024RoPE,
author = {Heo, Byeongho and Park, Song and Han, Dongyoon and Yun, Sangdoo},
title = {Rotary Position Embedding for Vision Transformer},
year = {2024},
isbn = {978-3-031-72683-5},
publisher = {Springer-Verlag},
address = {Berlin, Heidelberg},
url = {https://doi.org/10.1007/978-3-031-72684-2_17},
doi = {10.1007/978-3-031-72684-2_17},
booktitle = {Computer Vision – ECCV 2024: 18th European Conference, Milan, Italy, September 29–October 4, 2024, Proceedings, Part X},
pages = {289–305},
numpages = {17},
location = {Milan, Italy}
}

@article{Mao2021TowardsRV,
  title={Towards Robust Vision Transformer},
  author={Xiaofeng Mao and Gege Qi and Yuefeng Chen and Xiaodan Li and Ranjie Duan and Shaokai Ye and Yuan He and Hui Xue},
  journal={2022 IEEE/CVF Conference on Computer Vision and Pattern Recognition (CVPR)},
  year={2021},
  pages={12032-12041},
  url={https://api.semanticscholar.org/CorpusID:235211752}
}

@inproceedings{5206848,
  author={Deng, Jia and Dong, Wei and Socher, Richard and Li, Li-Jia and Kai Li and Li Fei-Fei},
  booktitle={2009 IEEE Conference on Computer Vision and Pattern Recognition}, 
  title={ImageNet: A large-scale hierarchical image database}, 
  year={2009},
  volume={},
  number={},
  pages={248-255},
  doi={10.1109/CVPR.2009.5206848}}

\newpage
\appendix

\section{Experimental Setup and Hyperparameters}

  \begin{table}[H]
    \centering
    \caption{Model architecture and training hyperparameters used in all experiments.}
    \label{tab:hyperparams}
    \begin{tabular}{l l}
    \toprule
    \textbf{Parameter} & \textbf{Value} \\
    \midrule
    \multicolumn{2}{c}{\textit{Input \& Tokenization}} \\
    \midrule
    Input resolution & $224 \times 224$ \\
    Patch size & $16 \times 16$ \\
    Number of patches & 196 \\
    Input channels ($C$) & 3 \\
    \midrule
    \multicolumn{2}{c}{\textit{ViT Architecture}} \\
    \midrule
    Embedding dimension ($D$) & 384 \\
    Number of encoder layers & 12 \\
    Number of attention heads & 8 \\
    Key/query dimension ($d_k$) & 48 \\
    Dropout (embedding) & 0.15 \\
    Dropout (attention) & 0.15 \\
    Dropout (MLP) & 0.15 \\
    Stochastic depth rate & 0.2 \\
    \midrule
    \multicolumn{2}{c}{\textit{Training Hyperparameters}} \\
    \midrule
    Batch size & 128 \\
    Optimizer & Adam \\
    Learning rate & $1 \times 10^{-3}$ \\
    Weight decay & $5 \times 10^{-2}$ \\
    Adam $\beta_1$ & 0.9 \\
    Adam $\beta_2$ & 0.999 \\
    Training epochs & 60 \\
    \bottomrule
    \end{tabular}
    \end{table}

\section{Additional Positional Encoding Scaling Results}
  \subsection{Full Learnable APE Scaling Results}
    \begin{table}[H]
      \centering
      \small
      \begin{tabular}{c|cc|cc}
      \toprule
      \textbf{$\alpha$} & \multicolumn{2}{c|}{\textbf{$\Delta$SSDC (RPI)}} & \multicolumn{2}{c}{\textbf{Fragility Score}} \\
      & Mean & Std & Mean & Std \\
      \midrule
      1.0 & 0.4725 & 0.0228 & 0.4338 & 0.0127 \\
      0.9 & 0.3850 & 0.0301 & 0.4520 & 0.0135 \\
      0.8 & 0.3125 & 0.0259 & 0.4780 & 0.0146 \\
      0.7 & 0.1845 & 0.0342 & 0.5145 & 0.0145 \\
      0.6 & 0.0975 & 0.0109 & 0.5597 & 0.0169 \\
      0.5 & 0.0475 & 0.0083 & 0.5975 & 0.0202 \\
      0.4 & 0.0000 & 0.0000 & 0.6272 & 0.0189 \\
      0.3 & 0.0000 & 0.0000 & 0.6425 & 0.0171 \\
      0.2 & 0.0000 & 0.0000 & 0.6483 & 0.0210 \\
      0.1 & 0.0000 & 0.0000 & 0.6522 & 0.0245 \\
      \bottomrule
      \end{tabular}
      \caption{
      Effect of positional embedding magnitude $\alpha$ on index-based spatial organization and robustness.
      $\Delta$SSDC (measured under RPI) captures the recovery of index-based spatial structure after the first encoder block.
      As $\alpha$ decreases, $\Delta$SSDC collapses, indicating the breakdown of index-based spatial organization, while fragility increases sharply in the same regime before plateauing once spatial structure is lost.
      }
      \label{tab:pe_scaling_full}
    \end{table}

    These results exhibit a clear three-regime structure. For large values of $\alpha$, $\Delta$SSDC remains high and fragility is relatively low, indicating stable index-aligned spatial organization. As $\alpha$ decreases past a critical range (around $\alpha \approx 0.5$–$0.6$), $\Delta$SSDC collapses sharply to zero, coinciding with a rapid increase in fragility. Below this threshold, fragility plateaus despite further reductions in $\alpha$, suggesting that once index-aligned spatial structure is lost, additional degradation of positional signal has limited further impact on robustness.

  \subsection{Sinusoidal Positional Encoding Scaling Results}
    \begin{table}[H]
      \centering
      \small
      \begin{tabular}{c|c|c}
      \toprule
      \textbf{$\alpha$} & \textbf{$\Delta$SSDC (RPI)} & \textbf{Fragility Score} \\
      \midrule
      1.0 & 0.29  & 0.4415 \\
      0.9 & 0.28  & 0.4526 \\
      0.8 & 0.25  & 0.4674 \\
      0.7 & 0.24  & 0.4868 \\
      0.6 & 0.25  & 0.5065 \\
      0.5 & 0.21  & 0.5204 \\
      0.4 & 0.17  & 0.5492 \\
      0.3 & 0.13  & 0.5756 \\
      0.2 & 0.07  & 0.6091 \\
      0.1 & 0.022 & 0.6357 \\
      \bottomrule
      \end{tabular}
      \caption{
      Effect of sinusoidal positional embedding magnitude $\alpha$ on index-based spatial organization and robustness (single-seed experiment).
      $\Delta$SSDC (measured under RPI) captures the recovery of index-based spatial structure after the first encoder block.
      We report no standard deviations because results are obtained from a single seed; this experiment is intended as a qualitative confirmation rather than a statistically rigorous estimate. Specifically, it verifies that the same qualitative relationship observed with absolute positional embeddings (namely, the collapse of $\Delta$SSDC and the increase in fragility as $\alpha$ decreases) also holds for sinusoidal positional encodings.
      }
      \label{tab:sinupe_scaling_single_seed}
    \end{table}

    A similar overall relationship is observed for sinusoidal positional encodings, with decreasing $\alpha$ leading to reduced $\Delta$SSDC and increased fragility. However, in contrast to learnable APE, the transition is more gradual: $\Delta$SSDC degrades smoothly across the full range of $\alpha$ without a sharp collapse threshold, and fragility increases correspondingly in a continuous manner. This suggests that while the coupling between spatial structure and robustness persists across encoding types, the dynamics of how spatial organization degrades are encoding-dependent.

\section{Additional Representational Analyses}
  \subsection{Depth-wise Evolution of Spatial Structure Across Encoding Schemes}

    \begin{figure}[H]
      \centering
      \includegraphics[width=0.6\linewidth]{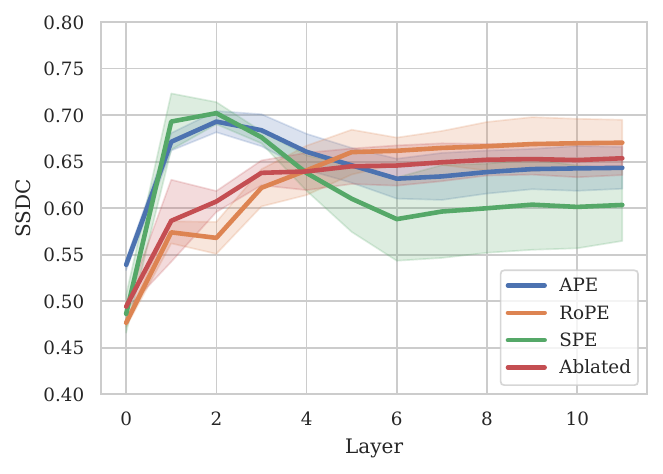}
      \caption{Layer-wise evolution of SSDC across model conditions in the unpermuted setting,
      averaged over 4 random seeds. Shaded regions indicate $\pm 1$ standard deviation.
      Distinct depth-wise trajectories emerge: APE and sinusoidal (SPE) models exhibit an
      early injection-like peak followed by gradual decay, while RoPE and ablated models
      show a more gradual increase across depth.}
      \label{fig:ssdc_depthwise_full}
    \end{figure}

    Figure~\ref{fig:ssdc_depthwise_full} shows the layer-wise evolution of SSDC across model
    conditions in the unpermuted setting. While all trained models develop non-trivial spatial
    structure, their depth-wise trajectories differ qualitatively.

    APE and sinusoidal (SPE) models exhibit an \emph{injection-like} pattern: SSDC rises sharply
    in the first few layers (typically peaking around layers 1--2), followed by a gradual decay
    or stabilization at deeper layers. This behavior is consistent with positional information
    being introduced additively at the input and propagated through the network, leading to a
    strong early-layer imprint that weakens with depth.

    In contrast, RoPE models display a more gradual increase in SSDC across depth, without a
    pronounced early peak. This aligns with the behavior observed under RPI in Section~5.3,
    and is consistent with positional information being incorporated multiplicatively within
    attention, allowing spatial structure to accumulate progressively across layers.

    A similar gradual trajectory is observed in ablated models. In the absence of positional
    signals, spatial structure must be inferred from patch content, making it intuitive that
    such structure is constructed incrementally over depth rather than injected early.

    These results highlight that while all models develop spatial structure, the \emph{mechanism}
    by which this structure emerges differs substantially across encoding schemes. Importantly,
    because SSDC in the unpermuted setting reflects both content-based and index-based effects,
    these trajectories alone do not distinguish the underlying spatial organization strategy;
    the RPI-based analysis in Section~5.3 is required for that separation.

  \subsection{Spatial Structure in Random Permutation Training (RPT)}

    \begin{figure}[H]
      \centering
      \includegraphics[width=0.6\linewidth]{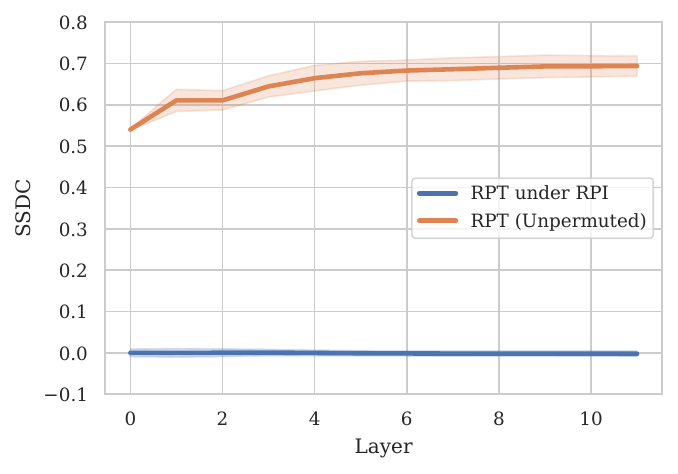}
      \caption{Layer-wise evolution of SSDC in RPT models. Without permutation at inference,
      SSDC increases gradually with depth, resembling the trajectory observed in ablated models.
      Under RPI, SSDC collapses to zero across all layers, indicating the absence of index-based
      spatial organization despite the presence of positional embeddings during training.}
      \label{fig:ssdc_rpt}
    \end{figure}

    Figure~\ref{fig:ssdc_rpt} shows the layer-wise evolution of SSDC for models trained under
    Random Permutation Training (RPT). We evaluate SSDC both in the standard (unpermuted)
    setting and under Random Permutation at Inference (RPI).

    In the absence of permutation at inference, RPT models exhibit a gradual increase in SSDC
    across depth, closely resembling the behavior observed in fully ablated models. This
    indicates that non-trivial spatial correlations can still emerge during training even when
    the mapping between token indices and spatial locations is randomized at every step.

    However, under RPI, SSDC collapses to approximately zero across all layers. This mirrors
    the behavior of ablated models and contrasts sharply with models trained with consistent
    positional embeddings, which retain substantial SSDC under permutation (Section~5.3).
    The absence of SSDC recovery indicates that RPT models do not develop an index-based
    spatial organization, despite the architectural presence of positional embeddings.

    Taken together, these results suggest that a stable mapping between token indices and
    spatial locations during training is necessary for index-based spatial organization to
    emerge. When this mapping is disrupted, as in RPT, spatial correlations that arise during
    training do not persist under permutation, and are therefore not anchored to token indices.

    We emphasize that while RPT models exhibit non-trivial SSDC in the unpermuted setting,
    this alone does not imply the presence of a well-formed spatial representation. Rather,
    it suggests that spatial correlations can arise from content statistics even in the absence
    of a consistent positional reference frame, consistent with the behavior observed in
    ablated models.

\section{Raw Accuracy and Fragility Metrics}
  \textbf{Overview.}
  We report raw top-1 accuracy on the clean ImageNet-100 validation set (\emph{normal\_acc}) and Fragility Scores (FS) under JPEG compression and Gaussian blur for all model conditions. All values are averaged over 4 random seeds with standard deviations.

  \begin{table}[H]
  \centering
  \small
  \begin{tabular}{l|cc|cc|cc}
  \toprule
  & \multicolumn{2}{c|}{\textbf{Normal Accuracy}} 
  & \multicolumn{2}{c|}{\textbf{JPEG FS}} 
  & \multicolumn{2}{c}{\textbf{Gaussian Blur FS}} \\
  \textbf{Model} & Mean & Std & Mean & Std & Mean & Std \\
  \midrule
  APE  & 0.6406 & 0.0018 & 0.4347 & 0.0132 & 0.1842 & 0.0141 \\
  SPE  & 0.6568 & 0.0069 & 0.4351 & 0.0086 & 0.2082 & 0.0054 \\
  RoPE & 0.6901 & 0.0042 & 0.2988 & 0.0065 & 0.1474 & 0.0131 \\
  Ablated & 0.5682 & 0.0056 & 0.6458 & 0.0119 & 0.2984 & 0.0085 \\
  RPT & 0.5595 & 0.0054 & 0.6297 & 0.0038 & 0.2427 & 0.0123 \\
  \bottomrule
  \end{tabular}
  \caption{
  Raw performance and fragility metrics across model conditions. Fragility Score (FS) is defined as the relative drop in accuracy under distribution shift. Higher values indicate greater sensitivity.
  }
  \label{tab:raw_metrics}
  \end{table}

  \textbf{Key Observations:}
  Across all conditions, models with positional encodings achieve higher clean accuracy than ablated and RPT models, with RoPE performing best overall. However, differences in clean accuracy (e.g., $\sim$69\% for RoPE vs.\ $\sim$64--66\% for APE/SPE) are modest compared to the much larger gaps observed in fragility.

  Under JPEG compression, a strong content-disrupting perturbation, models with positional encodings exhibit substantially lower fragility (APE/SPE $\sim$0.43, RoPE $\sim$0.30) than ablated and RPT models ($\sim$0.63--0.65). This large separation mirrors the presence or absence of index-based spatial organization identified in the main text.

  Under Gaussian blur, a milder perturbation, the same ordering is preserved but differences are compressed. RoPE remains the most robust, followed by APE/SPE, with RPT and ablated models exhibiting higher fragility. Notably, RPT consistently lies between PE-based and ablated models, suggesting that while it fails to develop a stable index-based spatial organization, it may still benefit from the presence of positional signals in a limited or indirect way.

  \textbf{Interpretation:}
  These results reinforce the central claim that robustness is primarily associated with the presence of a stable positional reference frame rather than raw accuracy. While positional encodings improve both accuracy and robustness, the magnitude of robustness differences far exceeds what would be expected from accuracy differences alone.

  At the same time, the non-identical behavior of RPT and ablated models (particularly under Gaussian blur) indicates that robustness cannot be explained solely by a binary distinction between index-based and non-index-based organization. Instead, these results suggest that positional signals may influence robustness through additional mechanisms (e.g., inductive biases on representation learning), even when they do not give rise to a stable index-based spatial structure.

  Overall, the raw metrics are consistent with, but do not by themselves establish, the mechanistic link proposed in the main text. This motivates the use of SSDC and permutation-based interventions to more directly probe the structure of representations.



\end{document}